\documentclass[conference]{IEEEtran}
\usepackage{cite}
\usepackage{url}
\newenvironment{promptblock}
{\begin{quote}\footnotesize\ttfamily}
{\end{quote}}
\usepackage{amsmath,amssymb,amsfonts}
\usepackage{algorithmic}
\usepackage{graphicx}
\usepackage{textcomp}
\usepackage{tabularx}
\usepackage{xcolor}
\usepackage{booktabs}
\usepackage{array}
\def\BibTeX{{\rm B\kern-.05em{\sc i\kern-.025em b}\kern-.08em
    T\kern-.1667em\lower.7ex\hbox{E}\kern-.125emX}}
\begin{document}

\title{Less Back-and-Forth: A Comparative Study of Structured Prompting}

\author{
\IEEEauthorblockN{Saurav Ghosh}
\IEEEauthorblockA{
\textit{Washington University in St. Louis} \\
Saint Louis, Missouri, USA \\
saurav.ghosh@wustl.edu
}
}

\maketitle

\begin{abstract}
Large language models (LLMs) are widely used for open-ended tasks, but underspecified prompts can lead to low-quality answers and additional interaction. This paper studies whether structured prompt design improves response quality while reducing user effort. We compare three prompt conditions: a raw prompt, a checklist-improved prompt, and a clarifying-question prompt. We evaluate these conditions across four task types---summarization, planning, explanation, and coding---using three LLM systems: ChatGPT, Claude, and Grok. Each output is scored with a unified rubric covering task completion, correctness, compliance, and clarity. Checklist-improved prompts achieved the highest mean rubric score, 7.50 out of 8, compared with 5.67 for raw prompts and 6.67 for clarifying-question prompts. Checklist prompts also produced the best quality-effort tradeoff, using fewer average tokens than both raw and clarifying prompts. These results suggest that a simple prompt checklist can improve LLM responses while reducing unnecessary interaction. 
\end{abstract}


\section{Introduction}

Large language models (LLMs) are now used for many everyday tasks, such as summarization, planning, explanation, and coding help. As these tools become more common, many users expect them to produce useful answers quickly. However, the quality of an LLM response often depends on the prompt. Prior work has shown that LLM behavior can change substantially depending on how instructions are written and presented to the model \cite{wei2022chain}. When a prompt is short, vague, or missing key details, the model may guess what the user wants. This can lead to weak answers, extra follow-up messages, and more effort before the result is good enough to use. 

This issue is central to human-AI interaction because the prompt is the primary mechanism through which users communicate goals, constraints, and expectations to the model. A better prompt can reduce ambiguity by telling the model what the task is, why it matters, and what kind of answer is needed. An even more interactive approach is to let the model ask clarifying questions before answering. In that case, the system does not simply guess missing details; it first tries to resolve them. From this view, prompt quality is not only a writing issue. It is also a coordination issue between a human and an AI system.

Both saving time and improving quality are important in this setting. A fast answer is not helpful if it is wrong, incomplete, or badly formatted. At the same time, a strong answer loses value if the user must spend too many turns revising the prompt or correcting the output. For real users, the best interaction is one that reaches an acceptable answer with as little extra effort as possible. This makes it important to study both final output quality and interaction effort together, rather than treating them as separate concerns.

Although many people informally claim that ``better prompts'' lead to better results, there is still a practical gap between advice and evidence. It is still unclear how much benefit comes from simply rewriting a prompt versus using an interactive clarification step, and whether those benefits remain similar across different kinds of tasks and different models. This paper addresses that gap through a small-scale comparative study of three prompt conditions: a raw prompt, a checklist-improved prompt, and a clarifying-question prompt. We evaluate these conditions across four task categories---summarization, planning, explanation, and coding---and across three LLM systems. We study three research questions:
\begin{itemize}
    \item[(1)] Do structured prompts improve output quality across task types and models?
    \item[(2)] What is the tradeoff among output quality, token usage, and interaction time?
    \item[(3)] Do structured prompts reduce interaction effort?
\end{itemize}
Together, these questions allow us to evaluate both final answer quality and the interaction cost required to obtain it. In this study, interaction cost is measured through turns-to-acceptance and token usage rather than wall-clock time. To address these questions, we make three contributions. First, we present a simple framework for comparing three common prompting strategies in a controlled way. Second, we test these strategies across multiple task types and models to examine whether the pattern is stable. Third, we evaluate both usefulness and effort by measuring answer quality, turns-to-acceptance, and token usage. 

\section{Related Work}
This section reviews prior work in three areas related to our study: prompt engineering and instruction-following, clarification in human-AI interaction, and evaluation of LLM output quality and user effort. We searched Semantic Scholar and PubMed, identified 50 papers, and included 21 papers after screening for relevance. 

\subsection{Prompt Engineering and Instruction-Following}
Prompt engineering studies how the wording, structure, and context of a prompt affect the quality of an LLM response. Studies show that prompts work better when they are clear, specific, and written with the right context in mind \cite{Chen2023Unleashing} \cite{Meskó2023Prompt} \cite{Cain2023Prompting} \cite{Anam2025Prompt} \cite{Sahoo2024A}. These findings suggest that prompt design is an important factor in response quality.

Researchers have studied both manual and automatic ways to improve prompts. Some findings show that automatic prompt design can match, and in some cases even outperform, prompts written by humans \cite{Zhou2022Large}. Work on instruction-tuned models also shows that they perform better when they learn from well-written prompts and clear instructions. This motivates our comparison between raw prompts and checklist-improved prompts, where the checklist makes the task goal, context, and expected answer format more explicit.

\subsection{Interactive Human-AI Interaction}

Working with LLMs often involves clarification over multiple turns. Users do not always get the right result from the first prompt. Instead, they often revise their requests after seeing the model's response or noticing that important details are missing \cite{Zamfirescu-Pereira2023Why} \cite{Desmond2024Exploring}. This makes clarification a normal part of human-AI interaction.

Prior work shows that interactive systems can help users refine prompts and reach better results more easily \cite{Mishra2025PromptAid:}. By helping users identify and fill in missing information, these systems can reduce effort and improve the overall interaction.

At the same time, prompt refinement is not always easy for non-expert users. Many users do not know what details to include, and some rely too much on human-to-human communication habits when writing prompts. This is important for our study because it supports the idea that clarification can be useful when the original prompt is incomplete.

\subsection{Output Quality and User Effort}

The quality of an LLM's output depends not only on the model itself, but also on how clear and well-structured the prompt is \cite{Tang2024Harnessing}. Studies show that stronger prompts often produce better feedback and more useful responses than weak or poorly written prompts \cite{Jacobsen2025The}. This supports the idea that prompt quality should be treated as an important factor when judging final output quality.

Recent work also shows that evaluation should look at more than correctness alone. Researchers now consider factors such as factuality, reasoning quality, readability, usability, user satisfaction, and time saved \cite{Nayab2024Concise}. Some studies further suggest that structured prompting can reduce the number of revision cycles and improve both efficiency and perceived output quality \cite{Kim2023EvalLM:}. This broader view is important for our study because we are interested in both output quality and user effort.

Although research in this area has made strong progress, several challenges remain. The value of prompt engineering is not always the same across models or tasks. Some advanced models appear to benefit less from more complex prompting, and there is still no single standard framework for evaluating prompt success \cite{Atreja2024Prompt}. Non-expert users also continue to face difficulties in writing effective prompts, especially when they want both high-quality outputs and low interaction effort.

\subsection{Research Gap and Challenges}

\begin{table*}[t]
\centering
\caption{Key claims and supporting evidence identified in the reviewed papers.}
\label{tab:claims-evidence}
\renewcommand{\arraystretch}{1.15}
\begin{tabular}{@{}p{6.2cm}p{9.0cm}p{1.8cm}@{}}
\toprule
\textbf{Claim} & \textbf{Reasoning} & \textbf{Papers} \\
\midrule
More structured prompts improve LLM output quality
& Consistent gains in accuracy and quality with well-designed prompts
& \cite{Chen2023Unleashing,Jacobsen2025The,Cain2023Prompting} \\

Better prompts reduce user effort/time spent
& Structured prompting reduces the number of revisions needed; users report higher efficiency
& \cite{Anam2025Prompt,Kim2023EvalLM:} \\

Automated prompt optimization can exceed human-crafted prompts
& Automatic methods outperform baselines on many tasks, although they may lack domain nuance
& \cite{Zhou2022Large} \\

Non-experts struggle without guidance
& Novices often use ad hoc revisions; training and guidance improve outcomes
& \cite{Knoth2024AI,Zamfirescu-Pereira2023Why} \\

Prompt effectiveness varies by model/task
& Some advanced models benefit less from complex prompting; task--model fit matters
& \cite{Wang2024Do} \\

No universal framework exists for evaluating prompt success
& Metrics vary widely; subjective and user-centered measures are still developing
& \cite{Mudrik2025Prompt} \\
\bottomrule
\end{tabular}
\end{table*}

These prior findings leave several important challenges open. Prompt effectiveness does not stay the same across all models and tasks. Some stronger models may gain less from complex prompting, and current findings may not fully transfer as LLMs continue to change \cite{Wang2024Do}. This makes it difficult to turn current results into a stable set of best practices.

Another challenge is accessibility. Writing effective prompts still takes skill, revision, and experience, which can make these methods harder for everyday users to apply consistently \cite{Knoth2024AI}. Automatic prompt-improvement tools may help, but they do not remove the need for human judgment, especially for tasks with subtle requirements or domain-specific details \cite{2024What}.

A further challenge is evaluation. Recent work has expanded beyond accuracy to consider broader factors such as readability and user satisfaction \cite{Mudrik2025Prompt}. However, there is still no widely accepted standard for measuring prompt quality or prompt success. These limitations motivate our study of response quality and interaction effort across prompt strategies.

\section{Methods}

We evaluated the following hypotheses across three prompt conditions, four task categories, and three LLM systems.

\begin{itemize}
    \item H1: Checklist-improved prompts will produce higher-quality outputs than raw prompts.
    
    \item H2: Clarifying-question prompts will produce higher-quality outputs than both raw prompts and checklist-improved prompts.
    
    \item H3: Checklist-improved prompts and clarifying-question prompts will reduce interaction effort compared with raw prompts.
    
\end{itemize}

The study uses a small-scale comparative design to compare common prompting strategies by measuring both final answer quality and the interaction required to reach an acceptable output.

\begin{figure}[t]
    \centering
    \includegraphics[width=\columnwidth]{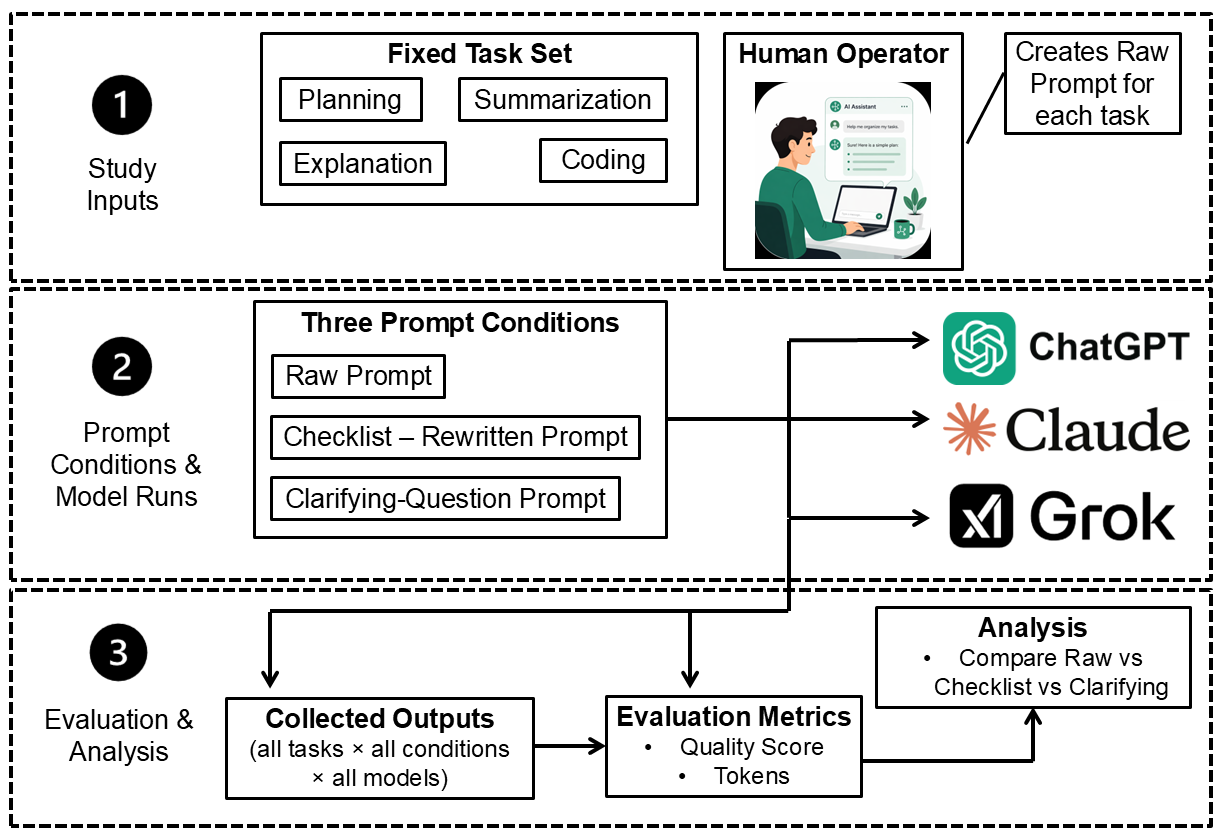}
    \caption{Study design overview. For each task, a raw prompt is evaluated alongside a checklist-improved prompt and a clarifying-question prompt. Each prompt condition is tested across multiple LLMs, and the resulting outputs are scored using the same rubric.}
    \label{fig:study-design}
\end{figure}

For each task, we first wrote a raw prompt. We then tested the same task using a checklist-improved prompt and a clarifying-question prompt, where the model asked 1--3 questions before producing the final answer. Each condition was run in a new session so that earlier context did not affect later results \cite{ouyang2022training}.

We evaluated three LLM systems: ChatGPT, Claude, and Grok, as shown in Figure~\ref{fig:study-design}. Each model was handled by one human operator to keep the data collection process organized: Gabriella Polach evaluated ChatGPT, Abdou Sow evaluated Claude, and Saurav Ghosh evaluated Grok. All operators followed the same task set, prompt conditions, and scoring rubric. For each model, we used the same version throughout the study and recorded the date and version used for each run. Token usage is measured separately for each model. We use the OpenAI tokenizer for ChatGPT, the Claude tokenizer for Claude, and the Lunary tokenizer for Grok. For every trial, we recorded input tokens and output tokens so that total token usage could be compared across conditions.

The study uses four task categories: summarization, planning, explanation, and coding. These categories were chosen because they represent common open-ended LLM use cases. We used the following raw prompts:
\begin{itemize}
    \item[(1)] ``Summarise this: [Abstract of \url{https://arxiv.org/abs/2201.11903}]''
    \item[(2)] ``Plan a vacation in Europe''
    \item[(3)] ``Explain this: [Abstract of \url{https://arxiv.org/abs/2201.11903}]''
    \item[(4)] ``Generate code for user input.''
\end{itemize}
The exact prompt templates used for each task and condition are provided in Appendix~\ref{appendix:prompt_templates}. Across conditions, the underlying task content was held fixed so that differences could be attributed to prompt structure rather than task changes.

We compared three prompt conditions. The first is Raw Prompt. This is the baseline condition. A human writes a short, direct prompt as a normal user might when first asking the model for help. The raw prompt is intentionally basic and may omit task-specific details. The second condition is the Checklist-Improved Prompt. In this condition, the raw prompt is rewritten using a short clarity checklist. The checklist includes three parts: 
\begin{itemize}
    \item \textbf{Roles/Rules}: specify what role the model should take or what limits it should follow.
    \item \textbf{Context}: explain who the output is for and why the task is being done.
    \item \textbf{Answer Format}: define how the final answer should be structured.
\end{itemize}
We chose these three parts because roles or rules define the expected behavior of the model, context reduces ambiguity about the task purpose, and the answer format makes the desired output structure explicit. These choices are also consistent with prior work showing that instruction-following improves with clearer prompts. For the Clarifying-Question Prompt, the model does not answer immediately. Instead, it first asks 1--3 clarifying questions. After the user answers those questions, the model gives its final response. This condition tests whether asking for missing information improves the final response, while also accounting for the extra interaction cost introduced by clarification. 

For each trial, we recorded the trial ID, task completion score, correctness score, compliance score, clarity score, total rubric score, interpretation label, input tokens, output tokens, and total turns to acceptance. For reference, we define a turn as one user message followed by one model reply. In the clarifying condition, the clarification exchange counts toward the total number of turns because it is part of the actual interaction cost. An output is marked as accepted when the evaluator decides it is good enough to use without further major revision. It is important to note that acceptance was based on the evaluator's judgment rather than an external user study. The interaction effort is measured through total turns to acceptance, input tokens, and output tokens. We use turns-to-acceptance as the main time-like measure. This measure captures the amount of back-and-forth required, not the actual elapsed time of the interaction. The primary outcomes in this study are output quality and interaction effort. Output quality is scored using four dimensions: task completion, correctness, compliance, and clarity. Each dimension receives a score from 0 to 2, producing a total rubric score from 0 to 8. The total rubric score is the main quality measure used in the aggregate, model-level, and task-level results. We then map the total score to a five-level interpretation: complete failure, poor, acceptable, strong, or flawless.

\begin{table}[t]
\centering
\caption{Unified quality scoring framework and final score mapping.}
\label{tab:unified_quality_framework}
\renewcommand{\arraystretch}{1.15}
\small
\begin{tabularx}{\columnwidth}{@{}>{\raggedright\arraybackslash}p{0.30\columnwidth}
                                c
                                >{\raggedright\arraybackslash}X@{}}
\toprule
\textbf{Dimension} & \textbf{Score} & \textbf{Meaning} \\
\midrule
Task Completion & 0 & Did not complete, or missed the main purpose. \\
                & 1 & Partly completed, but important parts are missing. \\
                & 2 & Fully completed as requested. \\
\addlinespace

Correctness & 0 & Contains major errors, or unreliable output. \\
            & 1 & Mostly correct, but has weak reasoning. \\
            & 2 & Correct, sound, and reliable. \\
\addlinespace

Compliance & 0 & Ignored the mentioned requirements. \\
           & 1 & Followed some, but missed important constraints. \\
           & 2 & Followed the stated constraints well. \\
\addlinespace

Clarity & 0 & Poorly structured, or needs major rewriting. \\
        & 1 & Understandable, but needs moderate editing. \\
        & 2 & Usable, and needs only minor or no editing. \\
\midrule

\textbf{Total Score} & \textbf{Final Score} & \textbf{Interpretation} \\
\midrule
0--1 & 1 & Complete failure \\
2--3 & 2 & Poor \\
4--5 & 3 & Acceptable \\
6--7 & 4 & Strong \\
8    & 5 & Flawless \\
\bottomrule
\end{tabularx}
\end{table}

For analysis, we compared the three prompt conditions pairwise: Raw vs.\ Checklist, Raw vs.\ Clarifying, and Checklist vs.\ Clarifying. For each comparison, we examined the mean and median quality scores, mean and median turns-to-acceptance, token usage, and acceptance patterns across tasks and models. The primary interpretation is that if a condition produces higher-quality results with fewer turns, that means it improves both result quality and interaction efficiency. If it improves quality but requires more turns, it may still be useful, though the benefit comes with an extra interaction cost.

As descriptive robustness checks, we compared whether the same pattern appeared across ChatGPT, Claude, and Grok. Consistent patterns across models would make the findings more convincing. We also compared results across task categories. This helps us see whether better prompting matters more for some task types, such as planning or coding, than for others. We did not conduct a full ablation of individual checklist components in this study. Instead, we treat such ablations as future work for testing whether role/rules, context, or answer format contributes most to the observed gains.

\section{Results}
The results show that checklist-improved prompts produced the strongest outputs overall, while raw prompts produced the lowest-quality outputs. Clarifying-question prompts improved over raw prompts in many cases, but they did not outperform checklist prompts when all results were combined.

\begin{table}[t]
\centering
\caption{Aggregate results by prompt condition. Scores are based on the 0--8 unified rubric.}
\label{tab:aggregate_results}
\normalsize
\resizebox{\columnwidth}{!}{%
\begin{tabular}{lcccc}
\hline
\textbf{Condition} & \textbf{Mean Score} & \textbf{Median Score} & \textbf{Mean Turns} & \textbf{Mean Tokens} \\
\hline
Raw & 5.67 & 6.00 & 1.00 & 962.25 \\
Checklist & 7.50 & 8.00 & 1.00 & 683.42 \\
Clarifying & 6.67 & 7.00 & 1.96 & 936.50 \\
\hline
\end{tabular}%
}
\end{table}

\begin{figure*}[t]
\centering
\begin{minipage}[t]{0.48\textwidth}
\centering
\includegraphics[width=\linewidth]{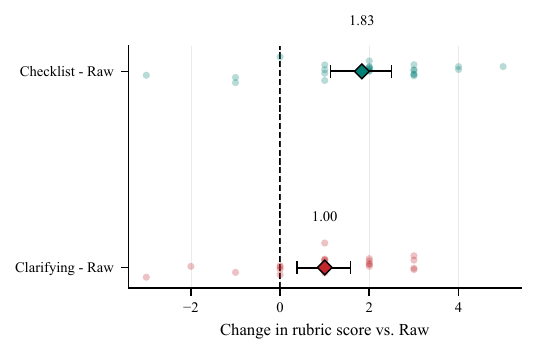}
{\normalsize (a) Change in rubric score relative to Raw.}
\end{minipage}
\hfill
\begin{minipage}[t]{0.48\textwidth}
\centering
\includegraphics[width=\linewidth]{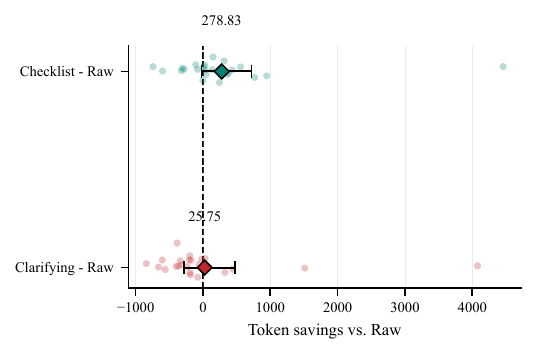}
{\normalsize (b) Token savings relative to Raw.}
\end{minipage}
\caption{Paired effects relative to the raw-prompt baseline. Each faint point represents one matched model-task trial, the diamond marks the mean paired change, and the horizontal interval shows a bootstrap 95\% confidence interval. Positive score changes indicate higher quality than Raw, while positive token savings indicate fewer tokens than Raw.}
\label{fig:paired_effects}
\end{figure*}

Table~\ref{tab:aggregate_results} shows that checklist prompts achieved the highest mean rubric score, 7.50 out of 8, while also using the fewest average tokens. Raw prompts had the lowest mean score, 5.67 out of 8, and the highest average token use. Clarifying prompts improved over raw prompts, with a mean score of 6.67 out of 8, but required almost twice as many turns on average.

These results support H1: checklist-improved prompts produced higher-quality outputs than raw prompts. H2 was only partially supported. Clarifying-question prompts improved over raw prompts, but they did not outperform checklist-improved prompts overall. H3 was supported for checklist prompts but not for clarifying-question prompts. Checklist prompts achieved stronger outputs in a single turn, whereas clarifying prompts usually required an additional question-answer exchange before the final response.

\begin{table}[t]
\centering
\caption{Mean rubric score by model and prompt condition.}
\label{tab:model_results}
\normalsize
\begin{tabular}{lccc}
\hline
\textbf{Model} & \textbf{Raw} & \textbf{Checklist} & \textbf{Clarifying} \\
\hline
ChatGPT & 5.62 & 7.88 & 7.12 \\
Claude & 6.00 & 7.25 & 7.38 \\
Grok & 5.38 & 7.38 & 5.50 \\
\hline
\end{tabular}
\end{table}

Table~\ref{tab:model_results} gives the model-level breakdown. ChatGPT and Grok achieved their highest mean scores with checklist prompts, while Claude achieved a slightly higher mean score with clarifying prompts. However, checklist prompts still required fewer turns overall. Grok showed the clearest gap between checklist and clarifying prompts; in several Grok trials, the clarifying questions were relevant, but the final answers sometimes strayed from the original task. Table~\ref{tab:task_results} shows that checklist prompts had the highest average score for planning, explanation, and coding. For summarization, checklist and clarifying prompts were tied.

\begin{table}[t]
\centering
\caption{Mean rubric score by task type and prompt condition.}
\label{tab:task_results}
\normalsize
\begin{tabular}{lccc}
\hline
\textbf{Task Type} & \textbf{Raw} & \textbf{Checklist} & \textbf{Clarifying} \\
\hline
Summarization & 6.00 & 7.67 & 7.67 \\
Explanation & 6.50 & 7.33 & 7.00 \\
Planning & 5.50 & 7.17 & 6.50 \\
Coding & 4.67 & 7.83 & 5.50 \\
\hline
\end{tabular}
\end{table}

The largest improvement appeared in coding. Raw coding prompts are often overly open to interpretation, whereas checklist prompts clearly specify the language, task, constraints, and expected output. This helped the models produce cleaner and more usable code. Planning also benefited from the checklist condition, as it added missing details such as budget, travel style, food preferences, and format.

The checklist condition gave the best tradeoff between quality and effort, with the highest average quality score and only one turn on average. The clarifying condition required almost two turns on average because the model first asked questions before giving the final answer. This extra step sometimes helped, but it also increased interaction time. Raw prompts also required one turn in our logged data. However, this should not be read as implying that raw prompts are equally efficient. In our experiment, raw prompts were not repeatedly revised after weak outputs. In real use, many raw outputs would likely require follow-up prompts because they were often incomplete, overly broad, or misaligned with the desired format. Token use should be interpreted carefully, as each model uses a different tokenizer. Therefore, token counts are most useful for comparing prompt conditions within this study rather than for making direct claims about absolute cost across providers. Checklist prompts used more input tokens than raw prompts because they included more instructions. However, they often reduced output length by making the expected answer more specific. As a result, checklist prompts had the lowest average total token count in our collected results.

For RQ1, structured prompts improved output quality across most models and task types, with the strongest gains for checklist prompts. For RQ2, checklist prompts showed the best balance between quality, token use, and interaction time. For RQ3, checklist prompts reduced interaction effort most clearly by producing strong outputs in one turn. Clarifying prompts may still help non-expert users who do not know what details to include, but in this study they added interaction cost and were less efficient than checklist prompts.

Overall, checklist prompting was the strongest strategy among the three conditions tested. It improved output quality, reduced unnecessary interaction, and produced more stable results across models and tasks. Even basic guidance about role or rules, context, and answer format can make the model's response more useful and easier to accept.

\section{Limitations and Future Work}

This study has several limitations. First, it evaluates a limited number of tasks, prompt conditions, and LLM systems. Therefore, the results should not be treated as a general conclusion for all prompting situations. Instead, they provide an early empirical comparison of how structured prompts may affect LLM output quality and user effort.

Second, the selected benchmark tasks may not fully transfer to more complex real-world or enterprise settings. Summarization, explanation, planning, and coding are common LLM tasks, but they do not cover all types of human-AI interaction.

Third, the quality scores are based on author evaluation using a rubric designed for this study. This helped keep scoring consistent, but it also introduces subjectivity because we did not use independent blind raters or report inter-rater reliability. The rubric also does not fully capture subjective factors such as user satisfaction, preferred writing style, or perceived usefulness. These factors may affect whether a real user accepts an LLM output.

Another limitation is that each LLM system was handled by one human operator. This kept data collection organized, but it may also introduce operator-specific variation in prompting, scoring, or judgment. Future work should use shared scoring procedures or multiple independent evaluators across all models.

Future work should evaluate the same prompt conditions with external human participants, for example, through platforms such as Prolific, which would allow the study to measure user-centered judgments of output quality, effort, and satisfaction. This would test whether the patterns observed with author scoring also hold for independent users. Such a study would require Institutional Review Board approval before participant recruitment. Future work should also use blind pairwise ratings, where evaluators compare two answers for the same task without knowing which prompt condition produced each answer. Another useful direction is to test individual checklist components, such as role/rules, context, and answer format, one at a time. This would help identify which checklist components contribute most to quality improvement and effort reduction.

\section{Conclusion}

This paper compared raw prompts, checklist-improved prompts, and clarifying-question prompts across four task types and three LLM systems. Checklist-improved prompts produced the strongest overall performance, achieving the highest mean rubric score while requiring only one turn on average. Clarifying-question prompts improved over raw prompts, but their benefits were less consistent and usually required extra interaction. These findings suggest that simple prompt structure can help users obtain better LLM outputs with less back-and-forth.

\bibliographystyle{IEEEtran}
\bibliography{ieee}

\appendices

\section{Experimental Prompt Templates}
\label{appendix:prompt_templates}

This appendix reports the exact prompt templates used in each experimental condition.

\subsection{Summarization}

\subsubsection{Raw Prompt}
\begin{promptblock}
Summarise this.\\
{[}PAPER ABSTRACT HERE{]}
\end{promptblock}

\subsubsection{Checklist-Improved Prompt}
\begin{promptblock}
Summarise the abstract below for a smart non-expert CS student. Use simple language. Keep the summary under 100 words. Focus on the main idea and why it matters. Write the answer as one short paragraph only, with no bullet points or headings.\\
\\
{[}PAPER ABSTRACT HERE{]}
\end{promptblock}

\subsubsection{Clarifying-Question Prompt}
\begin{promptblock}
Before answering, ask me exactly 3 short clarifying questions so you can write a better summary of the abstract below. Do not work on the task until I reply.\\
\\
Task:\\
Summarize this.\\
{[}PAPER ABSTRACT HERE{]}
\end{promptblock}

\subsection{Explanation}

\subsubsection{Raw Prompt}
\begin{promptblock}
Explain this.\\
{[}PAPER ABSTRACT HERE{]}
\end{promptblock}

\subsubsection{Checklist-Improved Prompt}
\begin{promptblock}
Explain the abstract below for a first-year graduate student. Use simple but technically correct language. Focus on the main idea and why the chain-of-thought helps. Write the answer in 2 short paragraphs only, with no bullet points or headings.\\
\\
{[}PAPER ABSTRACT HERE{]}
\end{promptblock}

\subsubsection{Clarifying-Question Prompt}
\begin{promptblock}
Before answering, ask me exactly 3 short clarifying questions so you can write a better explanation of the abstract below. Do not work on the task until I reply.\\
\\
Task:\\
Explain this.\\
{[}PAPER ABSTRACT HERE{]}
\end{promptblock}

\subsection{Planning}

\subsubsection{Raw Prompt}
\begin{promptblock}
Plan a vacation in Europe.
\end{promptblock}

\subsubsection{Checklist-Improved Prompt}
\begin{promptblock}
Plan a 7-day vacation in Europe for 2 adults. Keep the total budget around \$2500, excluding international flights. Focus on art, walkable cities, and vegetarian-friendly food. Avoid any plan that requires driving. Write the answer as a day-by-day itinerary and include a rough budget breakdown.
\end{promptblock}

\subsubsection{Clarifying-Question Prompt}
\begin{promptblock}
Before answering, ask me exactly 3 short clarifying questions so you can create a better travel plan. Do not work on the task until I reply.\\
\\
Task:\\
Plan a vacation in Europe.
\end{promptblock}

\subsection{Coding}

\subsubsection{Raw Prompt}
\begin{promptblock}
Generate code for user input.
\end{promptblock}

\subsubsection{Checklist-Improved Prompt}
\begin{promptblock}
Write Python code that prompts the user for a string. Check whether the string is a palindrome. Ignore spaces and letter case when checking. Print a clear result for the user. Write clean, runnable code only.
\end{promptblock}

\subsubsection{Clarifying-Question Prompt}
\begin{promptblock}
Before answering, ask me exactly 3 short clarifying questions so you can write better code for this task. Do not work on the task until I reply.\\
\\
Task:\\
Generate code for user input.
\end{promptblock}

\section{Trial-Level Evaluation}
\label{appendix:trial_level_results}

\begin{table}[t]
\centering
\caption{Compact trial-level evaluation summary. Scores are total rubric scores on the $0$--$8$ scale.}
\label{tab:trial_level_results}
\renewcommand{\arraystretch}{1.05}
\scriptsize
\begin{tabularx}{\columnwidth}{@{}>{\raggedright\arraybackslash}Xcccc@{}}
\toprule
\textbf{Trial ID} & \textbf{Score} & \textbf{Turns} & \textbf{\shortstack{Input\\Tokens}} & \textbf{\shortstack{Output\\Tokens}} \\
\midrule
Summarize1\_Grok\_Raw & 5 & 1 & 21 & 212 \\
Summarize1\_Grok\_Checklist & 7 & 1 & 74 & 146 \\
Summarize1\_Grok\_Clarifying & 7 & 2 & 144 & 148 \\
Explain1\_Grok\_Raw & 5 & 1 & 21 & 702 \\
Explain1\_Grok\_Checklist & 8 & 1 & 67 & 225 \\
Explain1\_Grok\_Clarifying & 6 & 2 & 139 & 814 \\
Plan1\_Grok\_Raw & 4 & 1 & 6 & 1472 \\
Plan1\_Grok\_Checklist & 8 & 1 & 67 & 1492 \\
Plan1\_Grok\_Clarifying & 3 & 2 & 95 & 1057 \\
Code1\_Grok\_Raw & 3 & 1 & 6 & 4676 \\
Code1\_Grok\_Checklist & 8 & 1 & 45 & 179 \\
Code1\_Grok\_Clarifying & 3 & 2 & 134 & 470 \\
Summarize2\_Grok\_Raw & 6 & 1 & 22 & 586 \\
Summarize2\_Grok\_Checklist & 7 & 1 & 58 & 175 \\
Summarize2\_Grok\_Clarifying & 7 & 2 & 36 & 956 \\
Explain2\_Grok\_Raw & 8 & 1 & 5 & 1253 \\
Explain2\_Grok\_Checklist & 5 & 1 & 46 & 943 \\
Explain2\_Grok\_Clarifying & 5 & 2 & 18 & 1318 \\
Plan2\_Grok\_Raw & 6 & 1 & 21 & 750 \\
Plan2\_Grok\_Checklist & 8 & 1 & 54 & 1318 \\
Plan2\_Grok\_Clarifying & 7 & 2 & 35 & 912 \\
Code2\_Grok\_Raw & 6 & 1 & 6 & 598 \\
Code2\_Grok\_Checklist & 8 & 1 & 50 & 862 \\
Code2\_Grok\_Clarifying & 6 & 2 & 19 & 1244 \\
Summarize1\_Claude\_Raw & 6 & 1 & 19 & 339 \\
Summarize1\_Claude\_Checklist & 8 & 1 & 49 & 629 \\
Summarize1\_Claude\_Clarifying & 8 & 2 & 125 & 1074 \\
Explain1\_Claude\_Raw & 6 & 1 & 17 & 778 \\
Explain1\_Claude\_Checklist & 8 & 1 & 64 & 416 \\
Explain1\_Claude\_Clarifying & 8 & 2 & 126 & 637 \\
Plan1\_Claude\_Raw & 6 & 1 & 5 & 548 \\
Plan1\_Claude\_Checklist & 5 & 1 & 48 & 260 \\
Plan1\_Claude\_Clarifying & 8 & 2 & 120 & 799 \\
Code1\_Claude\_Raw & 5 & 1 & 5 & 4874 \\
Code1\_Claude\_Checklist & 8 & 1 & 41 & 3891 \\
Code1\_Claude\_Clarifying & 8 & 2 & 107 & 3260 \\
Summarize2\_Claude\_Raw & 7 & 1 & 19 & 369 \\
Summarize2\_Claude\_Checklist & 8 & 1 & 54 & 311 \\
Summarize2\_Claude\_Clarifying & 8 & 2 & 36 & 370 \\
Explain2\_Claude\_Raw & 8 & 1 & 5 & 1165 \\
Explain2\_Claude\_Checklist & 7 & 1 & 48 & 354 \\
Explain2\_Claude\_Clarifying & 8 & 2 & 20 & 707 \\
Plan2\_Claude\_Raw & 7 & 1 & 18 & 607 \\
Plan2\_Claude\_Checklist & 7 & 1 & 65 & 411 \\
Plan2\_Claude\_Clarifying & 7 & 2 & 35 & 775 \\
Code2\_Claude\_Raw & 3 & 1 & 6 & 164 \\
Code2\_Claude\_Checklist & 7 & 1 & 51 & 860 \\
Code2\_Claude\_Clarifying & 4 & 2 & 21 & 353 \\
Summarize1\_ChatGPT\_Raw & 5 & 1 & 18 & 289 \\
Summarize1\_ChatGPT\_Checklist & 8 & 1 & 52 & 208 \\
Summarize1\_ChatGPT\_Clarifying & 8 & 2 & 43 & 824 \\
Plan1\_ChatGPT\_Raw & 5 & 1 & 5 & 578 \\
Plan1\_ChatGPT\_Checklist & 8 & 1 & 43 & 506 \\
Plan1\_ChatGPT\_Clarifying & 8 & 2 & 33 & 746 \\
Explain1\_ChatGPT\_Raw & 5 & 1 & 16 & 660 \\
Explain1\_ChatGPT\_Checklist & 8 & 1 & 48 & 270 \\
Explain1\_ChatGPT\_Clarifying & 7 & 2 & 41 & 972 \\
Code1\_ChatGPT\_Raw & 5 & 1 & 5 & 276 \\
Code1\_ChatGPT\_Checklist & 8 & 1 & 50 & 338 \\
Code1\_ChatGPT\_Clarifying & 8 & 2 & 35 & 849 \\
Summarize2\_ChatGPT\_Raw & 7 & 1 & 19 & 351 \\
Summarize2\_ChatGPT\_Checklist & 8 & 1 & 52 & 179 \\
Summarize2\_ChatGPT\_Clarifying & 8 & 2 & 33 & 305 \\
Explain2\_ChatGPT\_Raw & 7 & 1 & 5 & 519 \\
Explain2\_ChatGPT\_Checklist & 8 & 1 & 43 & 481 \\
Explain2\_ChatGPT\_Clarifying & 8 & 2 & 20 & 697 \\
Plan2\_ChatGPT\_Raw & 5 & 1 & 16 & 930 \\
Plan2\_ChatGPT\_Checklist & 7 & 1 & 56 & 332 \\
Plan2\_ChatGPT\_Clarifying & 6 & 1 & 31 & 1233 \\
Code2\_ChatGPT\_Raw & 6 & 1 & 6 & 106 \\
Code2\_ChatGPT\_Checklist & 8 & 1 & 50 & 341 \\
Code2\_ChatGPT\_Clarifying & 4 & 2 & 20 & 490 \\
\bottomrule
\end{tabularx}
\end{table}

\end{document}